\documentclass[journal,article,submit,moreauthors,pdftex]{mdpi}

\firstpage{1}
\pubvolume{xx}
\issuenum{1}
\articlenumber{5}
\pubyear{2019}
\copyrightyear{2019}
\history{Received: 24 January 2019; Accepted: data; Published: data}

\Title{DeepHMap++: Combined Projection Grouping and Correspondence Learning for Full DoF \linebreak Pose Estimation}

\Author{Mingliang Fu 
$^{1,2,3,}$*\href{https://orcid.org/0000-0002-6006-6699}{\orcidicon} and Weijia Zhou $^{1,2}$}

\AuthorNames{Mingliang Fu, Yuquan Leng, Haitao Luo and Weijia Zhou}

\address{%
$^{1}$ \quad State Key Laboratory of Robotics, Shenyang Institute of Automation, Chinese Academy of Sciences, Shenyang 110016, China\\
$^{2}$ \quad Institutes for Robotics and Intelligent Manufacturing, Chinese Academy of Sciences, \mbox{Shenyang 110016, China} \\
$^{3}$ \quad University of Chinese Academy of Sciences, Beijing 100049, China \\}

\corres{Correspondence: fumingliang@aliyun.com; Tel.: +xx-xx}

\abstract{In recent years, estimating the 6D pose of object instances with convolutional neural network (CNN) has received considerable attention. Depending on whether intermediate cues are used, the relevant literature can be roughly divided into two broad categories: direct methods and two-stage pipelines. For the latter, intermediate cues, such as 3D object coordinates, semantic keypoints, or virtual control points instead of pose parameters are regressed by CNN in the first stage. Object pose can then be solved by correspondence constraints constructed with these intermediate cues. In this paper, we focus on the postprocessing of a two-stage pipeline and propose to combine two learning concepts for estimating object pose under challenging scenes: projection grouping on one side, and correspondence learning on the other. We firstly employ a local-patch based method to predict projection heatmaps which denote the confidence distribution of projection of 3D bounding box’s corners. A projection grouping module is then proposed to remove redundant local maxima from each layer of heatmaps. Instead of directly feeding 2D--3D correspondences to the perspective-n-point (PnP) algorithm, multiple correspondence hypotheses are sampled from local maxima and its corresponding neighborhood and ranked by a correspondence--evaluation network. Finally, correspondences with higher confidence are selected to determine object pose. Extensive experiments on three public datasets demonstrate that the proposed framework outperforms several state of the art methods.}

\keyword{6D pose estimation; partial occlusion; projection grouping; correspondence evaluation}

\begin{document}

\section{Introduction} \label{sec1}
Estimating the full degree-of-freedom (DoF) pose of a rigid object, meaning 3D translation and 3D orientation from a single frame is an important topic in the realm of computer vision. A huge number of approaches have been proposed to address applications in the domain such as robotics, augmented reality and medical navigation \cite{ref1}. Real time is a key indicator for almost all applications from the three fields above. One line of solutions is modeling objects into a sparse set of feature points~ \cite{ref2}. For well-textured objects, this problem has been well addressed by constructing correspondence constraints between the prior model and scene image \cite{ref2,ref3}. However, both robustness and accuracy continue to be critical issues that limit existing methods under challenging scenarios \cite{ref4}. Thus, many researchers have recently begun to employ convolutional neural network (CNN) or ensemble learning~\cite{ref5} to address these issues.

Depending on whether intermediate cues such as 3D object coordinate \cite{ref6}, projection of virtual control points \cite{ref7}, or semantic keypoints \cite{ref8} are used, related approaches can be roughly divided into two categories: direct methods and two-stage pipelines. For a typical two-stage pipeline, intermediate cue is prepared in the first stage and pose parameters is computed by these intermediate cues in the back-end. Throughout the rest of this paper, the front-end and back-end correspond to the first and second phase of a two-stage pipeline, respectively. Instead of predicting directly the full DoF pose, Brachmann et al. \cite{ref6} first compute 3D object coordinates and the confidence map of scene pixels with random forests. Dense correspondences then are transferred instantly to a random sample consensus (RANSAC) based
optimization step. Encoding local feature of the input image makes 3D object coordinates inherently robust to partial occlusion and achieves top-level results on the Occluded LineMOD dataset \cite{ref6}. However, \mbox{Brachmann et al. \cite{ref6}} didn’t specially consider the case of symmetrical objects \cite{ref9}. Sparsity is another important requirement to ensure the robustness to heavy occlusion. In contrast to dense object coordinates \cite{ref6}, both projections of virtual control points~\cite{ref10} and bounding box’s corners \cite{ref7} are formally sparse. Oberweger et al. \cite{ref11} taken into account both sparsity and locality, and upgraded the robustness of BB8 \cite{ref7} (8 corners of the bounding box) by predicting projection heatmaps from random local patches. For simplicity, the method reported in Ref.~\cite{ref11} is named DeepHMap. As a part-based method, DeepHMap \cite{ref11} lifts the robustness to occlusion to a new level with simple local patches. However, predicted heatmaps always encounter multiple local maxima due to the absence of global information. Oberweger et al. \cite{ref11} directly select the global maxima to construct correspondence constraints without considering the rationality of corner projection. Compared with a wide range of intermediate cues, the postprocessing corresponding to the second stage is seldomly noticed. The studies \cite{ref12,ref13,ref14} are three of the few postprocessing stages that begin to divert attention to the back-end, all of which are less portable because of the depth of~customization.

\begin{figure}[H]
\centering
\includegraphics[width=7 cm]{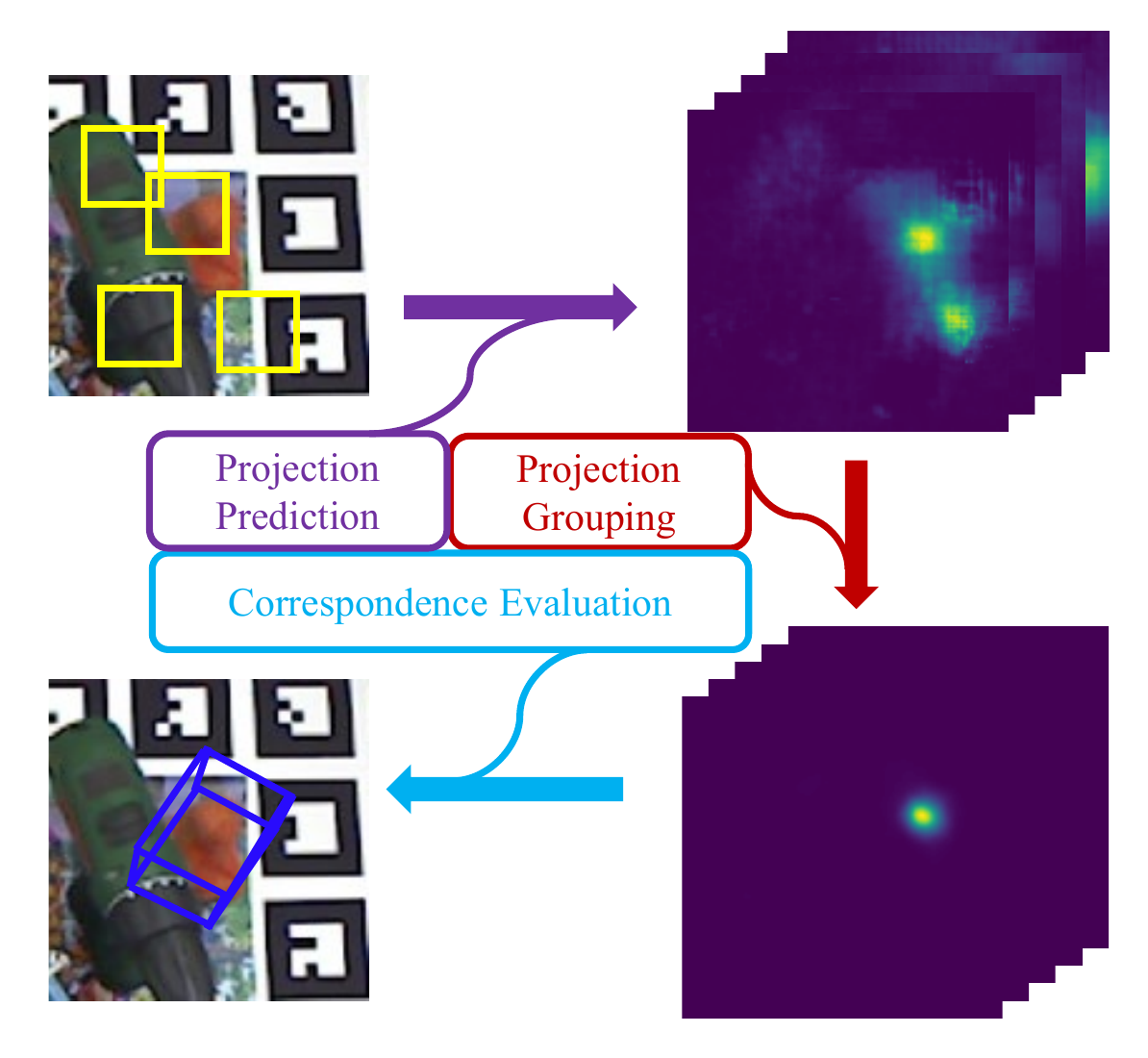}
\caption{Overview of the proposed two-stage approach for recovering the 6D pose. For any input patch (yellow box), its corresponding output from projection prediction module consists of projection heatmaps with eight layers.} \label{fig1}
\end{figure}

Motivated by the above analysis, we focus on the back-end of a two-stage pipeline for ensuring both accuracy and robustness in this paper. Given the simple yet efficient strategy of baseline \cite{ref11}, we~follow the same line to achieve projection heatmaps of 3D bounding boxes' corners (BBCs). For raw merged heatmaps from the baseline \cite{ref11}, a good postprocessing method should have the following features: (1) the postprocessing can be seamlessly integrated with the front-end network. That is, we do not have to spend extra effort to connect these two parts; (2) the back-end should be efficient enough, and introducing a heavy computational cost in exchange for improved accuracy is not advisable; and~(3)~unreasonable projection distribution on a single-layer of heatmaps should be properly excluded. To~this end, we present a two-stage approach as depicted in Figure \ref{fig1}. The proposed method consists of three parts, projection prediction, projection grouping and correspondence evaluation. A simple projection grouping module is designed firstly to learn spatial correlation of projection of different BBCs. Thus, unreasonable local maxima can be removed by geometric constraints learned with this projection grouping module and each layer of the filtered heatmaps contains only one peak. For~each layer of heatmap, each pixel stores the corresponding confidence of projection distribution. In fact, current projection predictions are still biased against the ground truth. Multiple correspondence hypotheses thus are sampled from both the only local maxima and its corresponding neighborhood instead of feeding directly 2D--3D correspondences to the perspective-n-point (PnP) method \cite{ref15}. Similar hypothesis sampling can be found in Ref. \cite{ref16}. Instead of random hypothesis selection or iterative refinement \cite{ref13}, we delegate all of these to a correspondence evaluation network \cite{ref17}. Finally, correspondences with higher confidence are chosen to calculate object pose. For brevity, we name our method DeepHMap++ in text below. 

In summary, the main contributions of this paper can be concluded as the following:
\begin{enumerate}[leftmargin=*,labelsep=4.9mm]
\item	We present a simple yet efficient projection grouping module for removing fake local maxima in each layer of projection heatmaps. The projection grouping module learns correlation constraints among projections of different BBCs and to select the optimal projection.
\item	In order to suppress different jitters during inference, multiple correspondence hypotheses are randomly sampled from local maxima and its corresponding neighborhood and ranked by a correspondence--evaluation network.
\item	We show the effectiveness of projection grouping and corresponding evaluation, and the corresponding two-stage pipeline achieves state-of-the-art performance on public benchmarks including LineMod dataset \cite{ref18}, Occluded LineMOD dataset \cite{ref6}, and YCB-Video dataset \cite{ref19}.
\end{enumerate}

The rest of the paper is structured as follows: an overview of the related works is provided in Section \ref{sec2}. Section \ref{sec3} describes the complete pipeline. Extensive evaluations and comparisons with several state-of-the-art baselines are demonstrated in Section \ref{sec4}. Conclusions and future work are presented in Section \ref{sec5}.

\section{Related Works} \label{sec2}
From the early feature-based approaches \cite{ref2,ref3} to the recent methods based on machine learning, the field of pose estimation for 3D rigid objects has accumulated a vast range of solutions. In this section, we confine ourselves to the field of 6D pose estimation in single frame via learning. As~mentioned in a previous section, related works on 6D pose estimation can be roughly divided into two groups: direct methods \cite{ref20,ref21,ref22,ref23,ref24,ref25,ref26,ref27,ref28,ref29,ref30,ref31,ref32} and two-stage pipelines \cite{ref4,ref6,ref7,ref8,ref10,ref11,ref13,ref14,ref19,ref33,ref34,ref35,ref36,ref37,ref38,ref39}.

\subsection{Direct Methods}
For direct methods, full DoF poses are directly encoded in both learning and inference. Attracted by the efficient LineMOD \cite{ref18} template, Tejani et al. \cite{ref20} proposed a latent-class based hough forest that employs a part-based version to improve the robustness to partial occlusion and clutter. Instead of random forest, a convolutional auto-encoder \cite{ref31} is trained from local view patches and generalizes well both on seen and unseen objects. A similar extension of local patch based regression can be found in Ref. \cite{ref32}. Wohlhart et al. \cite{ref21} presented a novel learning based descriptor mapping the object categories and viewpoints to Euclidean space. Thus, a large Euclidean distance between descriptors means different category attributes and distance of descriptors in Euclidean space is directly related to the difference between different views. More aggressively, a pose guided feature \cite{ref22,ref30} is designed to learn exact pose differences.

One-shot based 6D pose estimation has been recently frequently addressed in the literature. \linebreak In Ref. \cite{ref23}, a fully connected auto-encoder is employed to learn latent features. Significant progress in visual object recognition and detection has been made with deep learning. Therefore, many scholars have begun to predict pose parameters in one branch of deep neural networks. Following the state-of-the-art object detector, Mask R-CNN \cite{ref40}, a multi-task learning network \cite{ref27} with a pose branch is demonstrated. Similar end-to-end fashion for pose estimation can also be found in \linebreak Refs. \cite{ref26,ref28}. To reduce the reliance on data annotations, an implicit orientation learning \cite{ref29} is proposed via learning from samples processed by an augmented autoencoder. Mitash et al. \cite{ref25} presented a comprehensive framework for full DoF pose estimation via Monte Carlo tree search, which completely eliminates a time-consuming labeling step. More recently, a multi-view and multi-class framework \cite{ref24} demonstrates impressive 6D pose estimation via a multi-class representation of pose~space.

\subsection{Two-Stage Pipeline}
The biggest difference between direct method and two-stage pipeline is whether to use intermediate cues or not.  A common intermediate cue is segmented point cloud \cite{ref33} in a pick-and-place system. 6D pose of a rigid object instance is achieved by aligning the segmented point cloud with a pre-scanned 3D model. 3D object coordinate \cite{ref6} is another flexible intermediate cue, which has been proven to be very efficient for 6D pose estimation \cite{ref6,ref8,ref35,ref38,ref41} and camera localization \cite{ref16}. To cope with a multi-object case, 3D object coordinates together with object labels \cite{ref34} are jointly employed intermediate cues.

Holistic methods \cite{ref14,ref19,ref36} globally formulate the pose detection issue and feed directly the entire scene image into the regression network. What we want to highlight here is SSD-6D (SSD denotes single shot detector) \cite{ref36}, which originally
constructs a 6D hypothesis from 2D bounding boxes. In contrast to 3D object coordinates, a 2D bounding box is an implicit intermediate cue that doesn’t explicitly contain 3D information. After significant progress \cite{ref42} has been made in the face of conventional clutter scenes \cite{ref18}, researchers begin to shift their attention to more challenging scenes under severe occlusion \cite{ref6}. Compared with the part-based approach \cite{ref6}, a holistic method such as SSD-6D is more likely to be disturbed by the foreground occlusion when constructing the mapping to pose space.

Crivellaro et al. \cite{ref10} put a novel intermediate cue, virtual control points, into our view. 3D~pose of an object part is represented as projections of virtual control points, making it possible to handle poorly textured objects under partial occlusion and heavy clutter. To avoid manually selecting parts of a special object like Crivellaro et al. \cite{ref10}, more general-purpose virtual control points, BBCs \cite{ref7} are utilized to construct 2D--3D correspondences. Following the same principle of BB8 \cite{ref7}, \linebreak Oberweger et al. \cite{ref11} proposed training CNN from random local patches and achieved state-of-the-art performance. A similar part loss \cite{ref43} or part response \cite{ref44} based method also shows strong robustness to occlusion in other areas such as face detection. In multi-branch networks \cite{ref19,ref39}, segmentation mask, 2D~ bounding box, and object’s center are frequently employed as intermediate cues. Different from virtual control points mentioned above, semantic keypoints \cite{ref37} have also been proved to be an effective intermediate cue. Unfortunately, time-consuming auto-extracting of semantic keypoints hinders its real-time applications.

The following postprocessing is equally important after the acquisition of intermediate cues. After achieving 3D object coordinates using random forests in the first stage, a novel pose agent \cite{ref13} is designed to repeatedly refine pose hypotheses. This is the only reinforcement learning based example that we can find in the back-end. In the case of a differentiable RANSAC (DSAC) based pipeline \cite{ref16}, finite differences used in refinement gradients lead to high gradient variance during the end-to-end learning. To address the remaining issues in DSAC, a fully differentiable backend \cite{ref12} is proposed for camera localization. Compared with intermediate cues in the front-end, research about postprocessing is still relatively deficient.

\section{Methods} \label{sec3}
According to the definition of the preceding statement, our method belongs to a typical two-stage method. In the front-end, projection heatmaps of 3D BBCs are predicted by the tutorial described in DeepHMap \cite{ref11}. The core task of the paper is to design a comprehensive postprocessing in the back-end. Our proposed postprocessing consists of two modules: projection grouping and correspondence learning based hypothesis selection. We describe each necessary step in this section.

\subsection{Local Patch Based Heatmap Prediction}
DeepHMap \cite{ref11} uses an asymmetric hourglass network for predicting projection heatmaps, which takes a random local patch with size of $ 32\times32 $ as input and produces corresponding predicted heatmaps with size of $ 128\times128 $. Different from direct projection prediction of BBCs with a holistic patch in BB8 \cite{ref7}, DeepHMap outputs projection heatmaps that denote a confidence distribution of projection. Compared with projection heatmaps, direct pose regression is a more demanding task. Different predicted heatmaps from random local patches are then merged via simple averaging, which constantly produces multiple local maxima in single channel of heatmaps. A more flexible strategy instead of the global maxima \cite{ref11} is adopted and described in detail in the subsection below.

\subsection{Projection Grouping}
For DeepHMap \cite{ref11}, the key to improve the robustness to heavy occlusion is to feed random local patches instead of holistic objects of interest to CNN. However, local patches mean that the correlation between different parts of a special object is ignored. During the inference, each sample in the minibatch predicts projection heatmaps according to its own content. In each channel of the merged heatmaps, multiple local maxima can be frequently found. Oberweger et al. \cite{ref11} select the global maxima to eliminate this ambiguity. However, the global maxima is not always the optimal~choice.

To solve these ambiguities more thoroughly, we propose a simple projection grouping module to guide the projection selection. For projection distribution on a single channel of heatmaps, the~rationality of its location can be evaluated by constraints from two aspects: correlation constraints among projections on different channels on one side, and correspondence constraints between 3D BBCs and their corresponding 2D projections on the other. Next, we elaborate on the design process of the network architecture with considering correlation constraints, and correspondence constraints are fused in subsequent correspondence evaluation step. The first difficulty we have to face is that the number of local maxima on each channel of the heatmaps is always in dynamic change. Each channel of the merged heatmaps may contain projection clusters ranging in number from zero to many. For local patches from background or occlusion areas, heatmaps may not contain peaks. Before detailing more design details, we first revisit the strategy employed in DeepHMap. In particular, let $X = \left\{ {{l_1},{l_2},...,{l_8}} \right\}$ represents predicted heatmaps consisting of eight channels corresponding to different BBCs. In order to get a group of projections from different heatmap channels, \linebreak Oberweger et al. \cite{ref11} consistently choose the global maxima. This simple strategy can be written as:

\begin{equation}
\forall i,\;\;1 \le i \le 8,\;{y_i} = \max \left( {{l_i}} \right), \label{equ1}
\end{equation}
where $\max \left(\cdot \right)$ is a function that takes the global maxima from a single-channel heatmap, the output $ {y_i} $ denotes the $i$th channel of filtered heatmaps and contains the predicted projection of the corresponding BBC. Obviously, the above mentioned projection grouping described in Equation (\ref{equ1}) is carried out separately on a layer-by-layer fashion.

To fuse correlation constraints mentioned above, a simple fully connected network with residual architecture is employed to learn different projection cases. The task of projection grouping becomes constructing a mapping f with learned parameters $\psi $, such that

\begin{equation}
\forall i,\;\;1 \le i \le 8,\;\;{y_i} = f\left( {X,\psi } \right) + {l_i}. \label{equ2}
\end{equation}

\begin{figure}[H]
\centering
\includegraphics[width=12 cm]{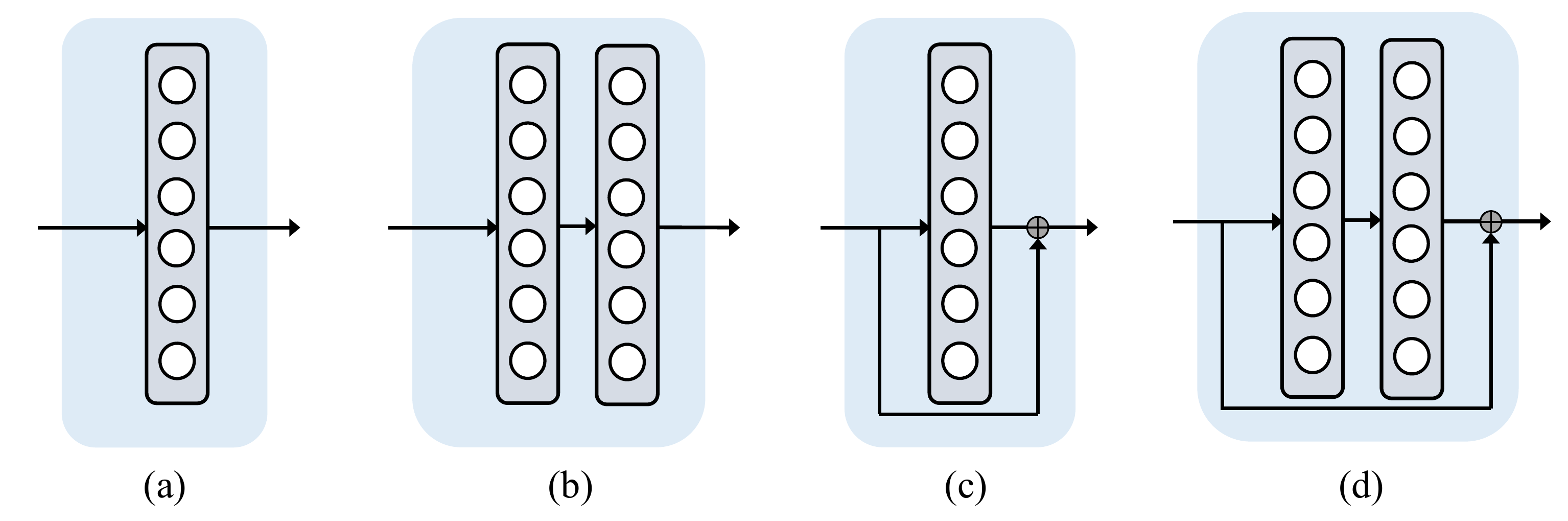}
\caption{Architecture of the projection grouping module. The projection grouping network adopts a residual structure \cite{ref46} consisting of a fully connected network based feedforward path and a shortcut connection (see case (\textbf{c}) and case (\textbf{d})). Compared with case (\textbf{c}), case (\textbf{d}) utilizes a fully connected two-layer feedforward network. The projection grouping layer takes merged heatmaps with a shape of $ [w, h, 8] $ as input and generates the same sized heatmaps containing only one peak in each channel. On the input side, merged heatmaps with a shape of $ [w, h, 8] $ are flattened to match the architecture of projection grouping module. On the output side, the immediate output is reshaped to generate filtered projection heatmaps. All layers have rectified linear unit (ReLU) \cite{ref47} 
activation function except for the output layer. Dropout layers \cite{ref48} are employed for the first dense layer, and the softmax function are placed after the add operation. Different configurations of projection grouping layer are detailed in the evaluation section. Additional case (\textbf{a}) and case (\textbf{b}) are plain module without shortcut connection, which are tested for comparison purposes.} \label{fig2}
\end{figure}

Compared with Equation (\ref{equ1}), the strategy given in Equation (\ref{equ2}) takes the correlation constraints among different channels into account. As shown in Figure \ref{fig2}, the projection grouping module takes merged heatmaps adjusted by the spatial transformation layer \cite{ref45} as input. For the input patch $ p_j $ of a minibatch consisting of $ N_{batch} $ batches, let ${Y_j} = \left[ {y_1^j,y_2^j,...,y_8^j} \right]$ and ${O_j} = \left[ {o_1^j,o_2^j,...,o_8^j} \right]$ represent the predicted heatmaps and expected heatmaps, respectively. Note that expected heatmaps are normalized to ensure that the maxima on each channel is equal to 1. Following the tutorial reported in DeepHMap, ground truth heatmaps are generated by placing a 2D Gaussian distribution at the ground truth projection for each channel. More details can be seen in Figure \ref{fig2}. In our practice, the immediate output of the last layer of projection grouping module is a feature vector $ V $ with $ N_{out} $ dimension. We~flatten the ground truth heatmaps ${O_j}$ to construct corresponding probability labels. The standard cross entropy loss for training can be written as

\begin{equation}
{L_{pg}} = \frac{1}{{{N_{batch}}}}\sum\limits_{j = 1}^{{N_{batch}}} {\sum\limits_i^{{N_{out}}} {{H_j}\left( {{v_i},s\left( {{{\hat v}_i}} \right)} \right)} },  \label{equ3}
\end{equation}
where $ H\left(  \cdot  \right) $ represents the cross entropy, $ N_{batch} $ denotes the number of samples in a minibatch, $ s\left(  \cdot  \right) $ is the softmax function, $ {\hat v_i} $ and $ v_{i} $ is the $ i $th element of output vector $ V $ and its corresponding probability label, respectively. With projection grouping module, most unmatched local maxima are removed and projection clusters corresponding to the ground truth are reserved. We test different configuration parameters of projection grouping module (see Figure \ref{fig2}), and the corresponding results can be found in Section \ref{sec4}.

\subsection{Correspondence Learning Based Hypothesis Scoring} \label{sec33}
Usually, only one peak is reserved for each channel after raw predicted heatmaps travel through the projection grouping module. However, it is still insufficient to directly utilize this maximum to construct correspondence constraints. CNN is inevitably disturbed by some explicit bias from occlusion, background clutter and noise in inference. To minimize these perturbations, we construct a hypothesis pool and throw all 2D--3D correspondences into a correspondence--evaluation network. Correspondence hypotheses are assigned a confidence score and high-confidence correspondences are collected to calculate the object pose. The process of assigning confidence to correspondence hypotheses is referred to as hypothesis scoring.

\subsubsection{Generating Hypothesis Pool}
We first describe the construction process of the hypothesis pool. As mentioned earlier, only one projection cluster is usually reserved for each channel after projection grouping. Eight projections of interest centered on the global maxima with a radius of $R$ are first determined. We use such a projection of interest to accommodate the bias caused by jitters. A total of ${N_{ch}}$ correspondence hypotheses are randomly sampled from each projection of interest, including the one corresponding to the peak. Additionally, the sampled points need to have a higher confidence than the predefined threshold. These~ $8 {N_{ch}}$ correspondence hypotheses are then fed into the subsequent correspondence--evaluation~network.

\subsubsection{Learning with a Hybrid Loss}
\begin{figure}[H]
\centering
\includegraphics[width=13 cm]{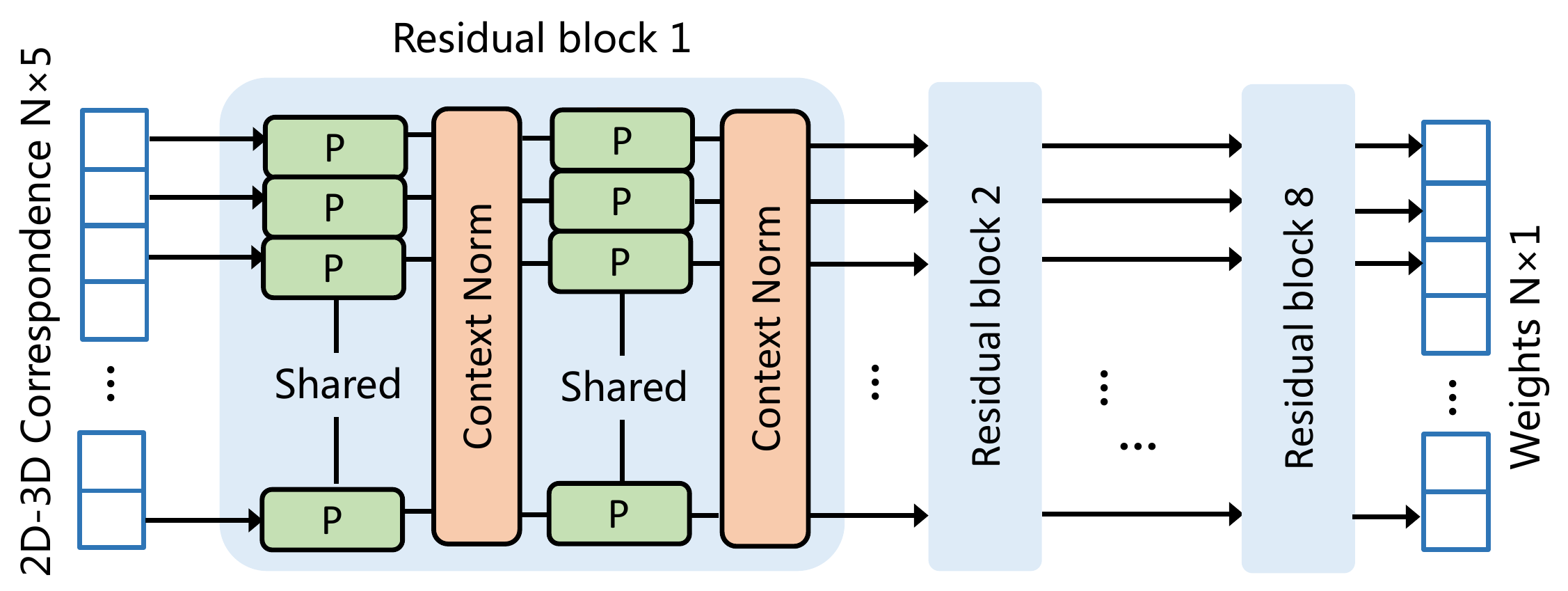}
\caption{Architecture of the correspondence--evaluation network. It takes 2D--3D correspondences as input and produces directly correspondence weights. The basic residual block consisting of weight-sharing perceptrons and context normalization. In practice, the multi-layer perceptrons are implemented using $ conv1d $ in tensorflow \cite{ref50}.} \label{fig3} 
\end{figure}

The hybrid loss of correspondence learning network \cite{ref17} consists of a classification term and a regression term. The input correspondences are assigned a weight that indicates whether they are inliers or outliers. Weighted correspondences are then utilized to formulate an essential matrix based regression loss. As shown in Figure \ref{fig3}, the correspondence--evaluation network in our case is formally similar to the correspondence-learning network \cite{ref17}. The input of our correspondence--evaluation network is 2D--3D correspondence $ {c_i} = \left[ {{p_i},{P_i}} \right] $ instead of keypoint pairs on stereo images. The loss function thus needs to be reformulated to accommodate the new input type. Let $  C = \left[ {{c_1},{c_2},...,{c_N}} \right] $ be a set of 2D--3D correspondences, where $ {p_i} = \left[ {{u_i},{v_i}} \right] $ is the predicted projection in the heatmaps and $ {P_i} = \left[ {{x_i},{y_i},{z_i}} \right] $ is the spatial coordinate of BBCs in the object coordinate system. For each object of interest, arbitrary pose can be represented by eight size-specific BBCs. Spatial coordinates of BBCs are thus reused when constructing different correspondences. For each 2D--3D correspondence, the~mapping takes the form of a $3 \times 4$ projection matrix:

\begin{equation}
\forall i,\;\;1 \le i \le N,\;\;{p_i} = H{P_i}.  \label{equ4}
\end{equation}

The vector $ {p_i} $ and $H{P_i}$ have the same direction, and Equation (\ref{equ4}) thus can be expressed in terms of a vector cross product:

\begin{equation}
{p_i} \times H{P_i} = 0.  \label{equ5}
\end{equation}

For the over-determined case that has more than six 2D--3D correspondences, the above \linebreak Equation (\ref{equ5}) can be rewritten in the following form:

\begin{equation}
{A_i}Vec\left( H \right) = 0,  \label{equ6}
\end{equation}
where $ {A_i} $ denotes the correspondence matrix
$\left[\begin{array}{*{20}{c}}
{{x_i}}&{{y_i}}&{{z_i}}&1&0&0&0&0&{-{u_i}{x_i}}&{-{u_i}{y_i}}&{-{u_i}{z_i}}\\
0&0&0&0&{{x_i}}&{{y_i}}&{{z_i}}&1&{-{v_i}{x_i}}&{-{v_i}{y_i}}&{-{v_i}{z_i}}\\
\end{array}\right. $
$\left. {\begin{array}{*{20}{c}}
{ - {u_i}}\\
{ - {v_i}}
\end{array}} \right]$, $ Vec\left( H \right) $ is the coefficient vector made up of entries from $H$. We now construct a $2N \times 12$ correspondence matrix $A$ by stacking Equation (\ref{equ6}) generated by each correspondence. The~projection matrix $H$ can be computed by performing the singular value decomposition (SVD) of $A$ and taking the unit singular vector corresponding to the smallest singular value \cite{ref15}. With classification term in the hybird loss, possible numerical instability \cite{ref17,ref49} in eigendecomposition are suppressed well. The singular value based regression term is replaced with common reprojection error:

\begin{equation}
{L_{geo}} = \frac{1}{{{N_h}}}\sum\limits_i^{{N_h}} {{{\left\| {H{P_i} - {p_i}} \right\|}_2}},   \label{equ7}
\end{equation}
where $ N_{h} $ represents the number of 2D--3D correspondences with a predicted label of 1. The~ classification term $ L_{cla} $ can be computed by a binary cross-entropy loss, which efficiently rejecting outliers with correspondence classification. Putting both the classification term and geometry term together, the overall loss of $N_{in}$ 2D--3D correspondences can be written as:

\begin{equation}
{L_{corr\_learn}} = \sum\limits_{i = 1}^{N_{in}} {\left( {\alpha {L_{cla}} + \beta {L_{geo}}} \right)}.  \label{equ8}
\end{equation}

The main differences between the correspondence--evaluation network we use here and the original case \cite{ref17} are detailed as follows: (1) Instead of 2D--2D correspondences obtained from stereo images, the network here takes 2D--3D correspondences as input and learns the mapping between projection heatmaps and BBCs. (2) Training loss of the network is reformulated by replacing the SVD based regression term to a general reprojection loss (see Equations (\ref{equ7}) and (\ref{equ8})). (3) Training dataset of the correspondence evaluation network is different from the conventional 2D case \cite{ref17} and 3D case~ \cite{ref51}. Additional correlation constraints are fused to imitate the projection distribution of 3D BBCs. More~training details are given in Section \ref {sec34}.

\subsection{Training Dataset} \label{sec34}
Our proposed two-stage pipeline can’t be trained via an end-to-end fashion because of non-differentiable paths connecting different modules. Three subtasks, prediction of projection heatmaps, projection grouping and correspondence evaluation thus are trained separately. In the first stage, a mixed dataset consisting of synthetic and real samples are generated according to the tutorial described in Ref. \cite{ref19}. The synthetic samples are collected by accumulating a series of discrete viewpoints, and the real parts are generated by segmenting the masked object of interest and then combining an additional in-plane rotation. This mixed dataset contains 200,000 samples, of which
the ratio of synthetic to real is 1 to 1. Hyper parameters of DeepHMap are completely preserved. Note~that DeepHMap is object-specific network and we also need to prepare similar object-specific training dataset for each~object.

Merged heatmaps are naturally collected to train the projection grouping module. As for the correspondence--evaluation network, it takes a set of 2D--3D correspondences as input. We thus synthesize a series of 2D--3D correspondences by projecting size-specific BBCs to the image coordinate system. Similar to the preparation of training dataset for DeepHMap, a sample set consisting of 200,000 
2D--3D correspondences is collected by placing a virtual camera at different viewpoints. Eight BBCs instead of a mesh model are placed at center of the view-sphere. Additional noises and outliers are added to augment synthesized samples. In practice, the weights in Equation (\ref{equ8}) are set to $ \alpha=1 $ and $ \beta = 0.15 $, respectively.

\section{Evaluations} \label{sec4}
\subsection{Datasets and Evaluation Metric}
In this section, three public datasets: LineMOD dataset \cite{ref18}, Occluded LineMOD dataset \cite{ref6} and YCB-Video dataset \cite{ref19} are employed to evaluate the proposed backend and integrated two-stage pipeline. The LineMOD dataset consists of 15 different object sequences and corresponding ground truth pose. The occluded version \cite{ref6} is generated by selecting images from LineMOD dataset, and these objects occlude each other to a large extent under different viewing directions. The YCB-Video dataset contains 21 different object sequences with significant image noise, illumination changes, background clutter and severe occlusion.

To evaluate the performance of pose estimation algorithms objectively, two popular metrics in this field, 2D reprojection error \cite{ref34} and AD{D|I} \cite{ref18} are employed to define a correctly estimated pose. With the 2D reprojection error, an estimated pose is accepted if the average reprojection error of all model points from the estimated pose and the ground truth pose is below five pixels. ADD
depicts a ratio between the average distance and the object’s diameter. ADI
is specifically designed to deal with symmetrical objects, of which the average distance is computed using the closest point of transformed model points. The default ratio in AD{D|I} is retained and set to 0.1.

\subsection{Architecture and Parameter Selection for Projection Grouping Module}
\begin{figure}[H]
\centering
\includegraphics[width=13 cm]{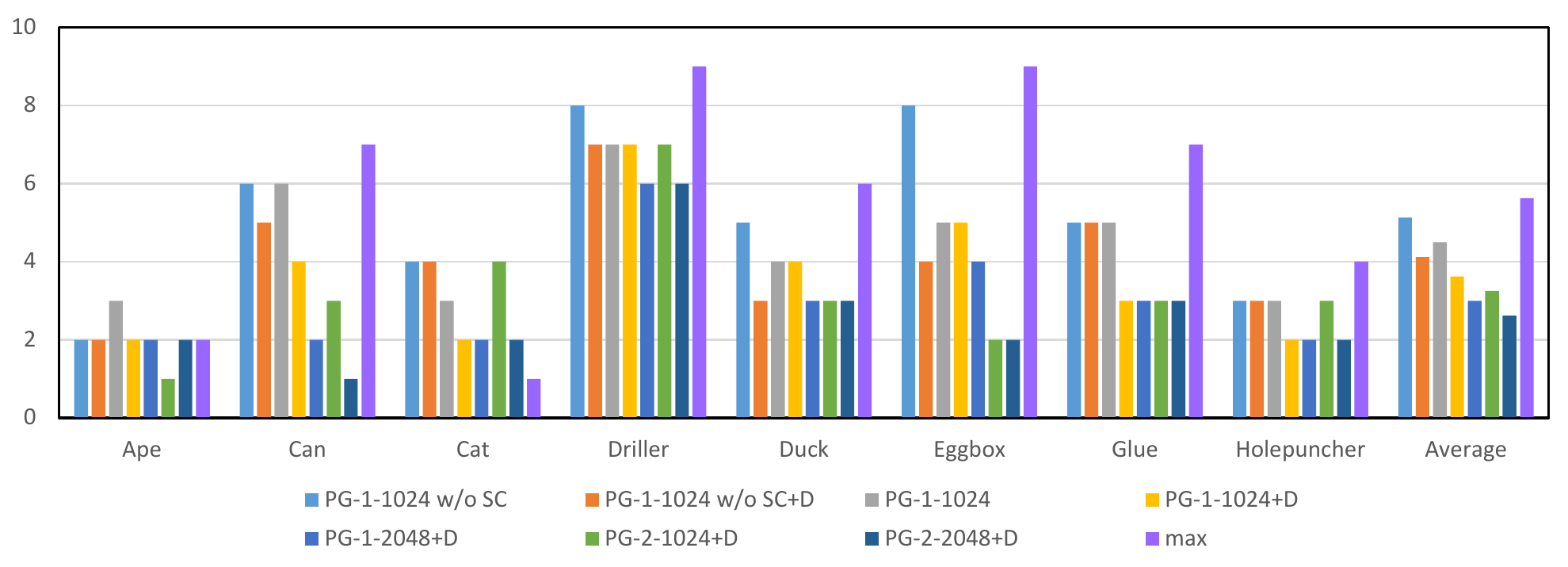}
\caption{Statistics of false projection selection per one hundred channels for different configurations on parts of the LineMOD dataset \cite{ref18}. A typical configuration can be expressed as PG-x-y w/o SC+D, where PG denotes the abbreviation of projection grouping module, w/o SC indicates that the network doesn’t contain shortcut connection (SC), D is the dropout layer, $x$ and $y$ represents the number of fully connected layers and dimensionality of the output space, respectively. max refers to the strategy utilized in DeepHMap. For results shown in figure, lower is better.} \label{fig4}
\end{figure}

\begin{figure}[H]
\centering
\includegraphics[width=13 cm]{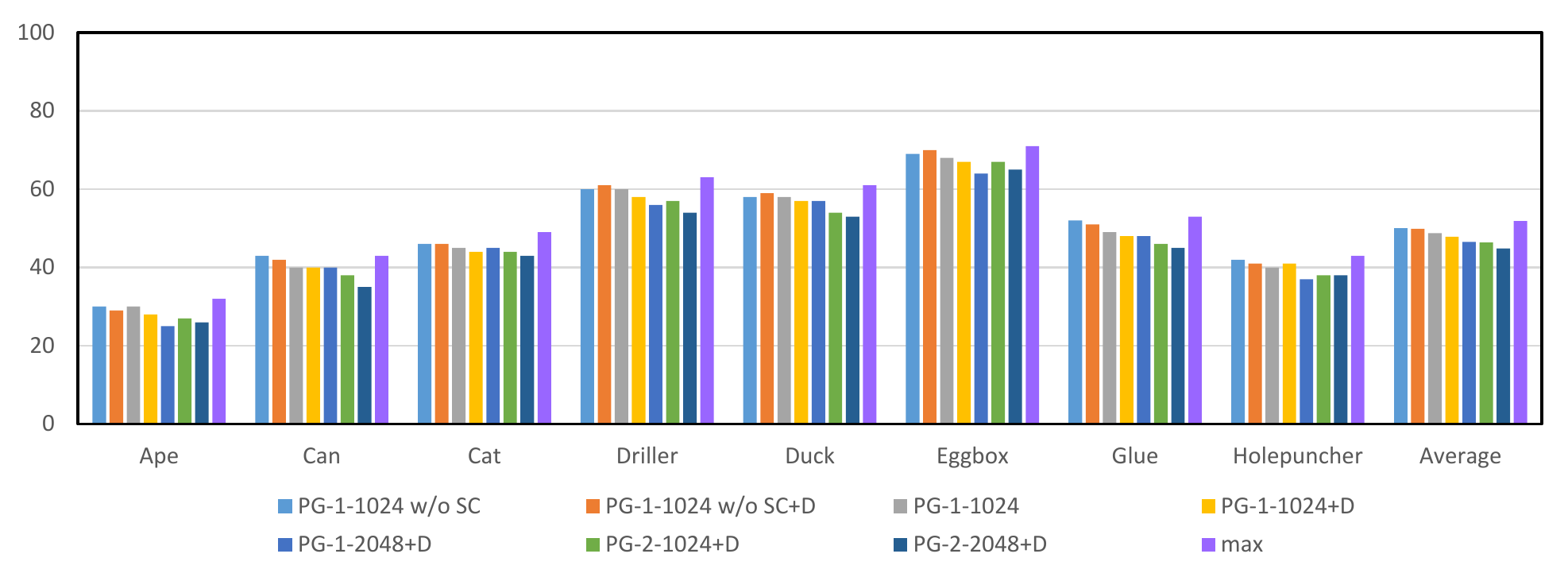}
\caption{Statistics of false projection selection per one hundred channels on Occluded LineMOD dataset \cite{ref6}. All test methods here are the same as Figure \ref{fig4}.} \label{fig5}
\end{figure}

Among the raw merged heatmaps from occluded scenes, multiple local peaks can frequently found in different channels. The one-size-fits-all rule in DeepHMap is not always able to find the optimal projection cluster, which defines a region of interest centered at the ground truth projection with a radius of 10 pixels. To quantify the effect of the projection grouping module, here we count the number of false projection selection (FPS) ${N_{ps}}$ per hundred channels. A projection selection is considered correct if its location is inside the corresponding projection cluster. The goal of our projection grouping module is to implicitly learn correlation constraints among projections of BBCs. To best meet the three design principles mentioned above, we test different configurations and report results in Figures \ref{fig4} and \ref{fig5}. The corresponding results provided by $ max $ 
function \cite{ref11} also have been included. Unless explicitly stated, results from DeepHMap don’t utilize feature mapping \cite{ref52}.

\begin{figure}[H]
\centering
\includegraphics[width=10.8 cm]{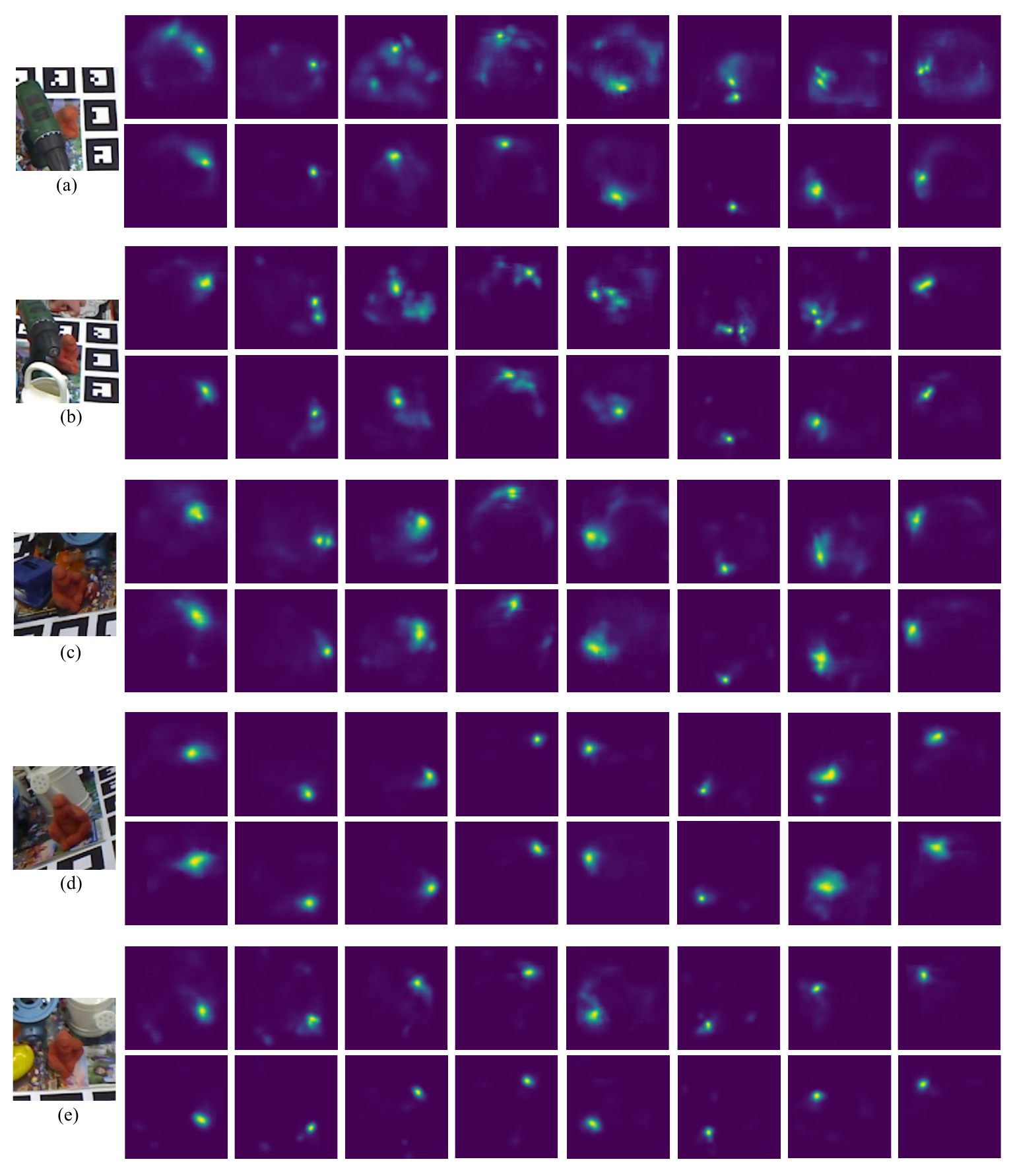}
\caption{Predicted projection heatmaps from different RGB
images of Occluded LineMOD dataset \cite{ref6}. From (\textbf{a}) to (\textbf{e}), the region of interest of different test frames and its corresponding predicted heatmap channels with DeepHMap (up) and projection grouping module (down) are given, respectively.} \label{fig6}
\end{figure}

As shown in Figures \ref{fig4} and \ref{fig5}, we observe the following results: (1) the projection grouping module with residual architecture and dropout layer achieves best results in FPS metric on both datasets; (2) Compared with the results corresponding to Occluded LineMOD dataset, all test methods give a lower number of FPS. It is in line with expectations because severe occlusions bring more interference to the inference of network; (3) In addition to the optimal configuration, other configurations of projection grouping module also outperform the $ max $ function in the baseline \cite{ref11}. The above experiments consistently demonstrate the effectiveness of projection grouping module. Such improvements can also be seen in Figure \ref{fig6}. For occluded cases, that is (Figure \ref{fig6}a,b) projection heatmaps from DeepHMap++ are much cleaner. As for non-occluded cases, improvements from projection grouping module are visually limited.

It benefits by learned correlation constraints and projection grouping module can find the matching projection that doesn't correspond to the global maxima (as shown in Figure \ref{fig6}). For the rest of the evaluation, we use the optimal configuration, that is, PG-2-2048+D.

\subsection{Correspondence Evaluation}
\begin{figure}[H]
\centering
\includegraphics[width=9 cm]{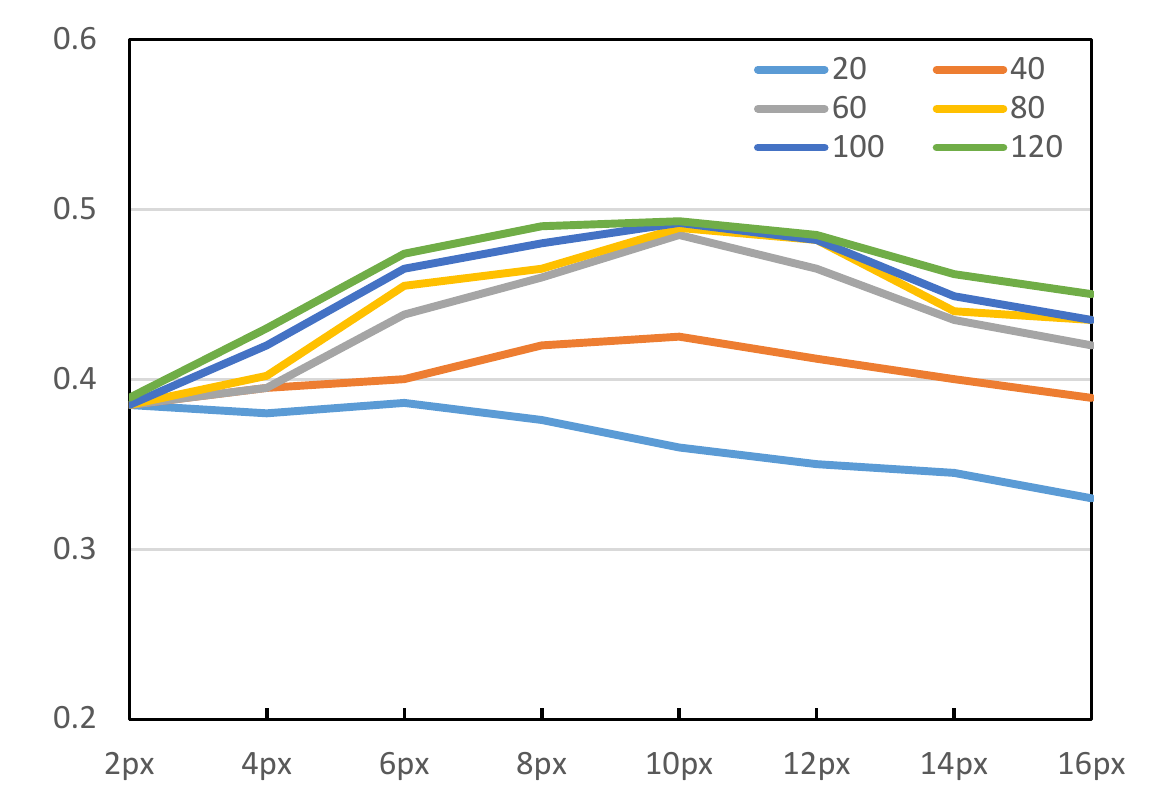}
\caption{Evaluation of our proposal with a varying radius of projection of interest (POI) and different sampled correspondences. A horizontal axis denotes the radius of POI in pixels. The vertical axis denotes the fraction of correctly estimated scenes under the 2D reprojection error metric.} \label{fig7}
\end{figure}

For the correspondence evaluation module, we evaluate it from three different perspectives. First, we evaluate the performance of correspondence evaluation network with varying sizes of projections of interest and varying numbers of sampled correspondences from each channel (see~Figure~\ref{fig7}). As~mentioned in Sections \ref{sec33} and \ref{sec34}, correlation constraints are employed to guide the learning of correspondence evaluation network. To verify the effectiveness of correlation constraints among the training dataset, we test two versions of the correspondence--evaluation networks: trained from dataset with correlation constraints (CorrNet) and without correlation constraints (CorrNet w/o CC). Here, correlation constraints are evaluated as the second factor. For the non-constraint case, we follow the similar procedure of a view-sphere based method \cite{ref18}. For each viewpoint, we randomize the 2D position of projections and achieve the corresponding 3D reference points by a back-projection function. Third, we evaluate the correspondence--evaluation module against a RANSAC 
based strategy employed in DeepHMap.

Regarding the radius of POI, Figure \ref{fig7} shows that
the increase of precision tends to saturation at 10 $ px $. As for the number of sampled correspondences, we can find that increasing from 60 to 80 or more only slightly affects the performance of the correspondence evaluation network. For the rest of the evaluation, we thus use the parametric values of $ r = 10~ px, n = 60 $.

With utilizing the identified radius of POI and the number of sampled correspondences, we then begin to evaluate the impact of correlation constraints on the CorrNet. It should be noted that the evaluation of DeepHMap on the LineMOD dataset hasn't been given, and thus we list the corresponding results from BB8 \cite{ref8} as a substitute. Table \ref{tab1} shows that both of the two versions significantly outperform BB8 \cite{ref11} on the LineMOD dataset \cite{ref18}. This is mainly because of the specially designed network for weighted correspondence and projection grouping, and boosting from DeepHMap. In the case of without correlation constraints, the average accuracy of CorrNet w/o CC is about $ 9.7\% $ higher than BB8 \cite{ref11} in AD{D|I} metric, and about $ 5.3\% $ higher in the 2D reprojection error metric.  In addition, correlation constraints among the training dataset can further improve the average accuracy of CorrNet w/o CC by $ 1.8\% $ under AD{D|I} metric, and about $ 0.7\% $ in a 2D-reprojection error metric. It proves the validity of correlation constraints on the correspondence selection.

Similar test results from Occluded LineMOD dataset \cite{ref6} can be found in Table \ref{tab2}. Margins between RANSAC based strategy in DeepHMap and two versions of CorrNet reach an average of  $ 1.6\% $ and $ 1.0\% $ in AD{D|I} metric, $ 5.2\% $ and $ 3.0\% $ in the 2D reprojection error metric. It confirms once again that correlation constraints can further improve the performance of CorrNet. CorrNet has a stronger ability to handle fake correspondences than the RANSAC based strategy \cite{ref11}.

\begin{table}[H]
\caption{Pose estimation results of two different versions: CorrNet and CorrNet w/o CC on the LineMOD dataset \cite{ref18}. The best results for each term are shown in bold.}
\centering
\begin{tabular}{l*{7}{c}}
\toprule
\multirow{2}*{\textbf{Sequence}\vspace{-4pt}} & \multicolumn{3}{c}{\textbf{AD{D|I}}} & \multicolumn{3}{c}{\textbf{2D Reprojection Error}}  \\
\cmidrule{2-7}
& \textbf{BB8 \cite{ref8}} & \textbf{CorrNet w/o CC} & \textbf{CorrNet}& \textbf{BB8 \cite{ref8}} & \textbf{CorrNet w/o CC} & \textbf{CorrNet}\\
\midrule
Ape	&	40.4 	&	49.7 	&	\textbf{51.2} 	&	96.6 	&	97.7 	&	\textbf{98.2} 	\\
Benchvise	&	91.8 	&	93.0 	&	\textbf{93.5} 	&	90.1 	&	97.5 	&	\textbf{98.1} 	\\
Camera	&	55.7 	&	61.7 	&	\textbf{62.9} 	&	86.0 	&	96.2 	&	\textbf{96.5} 	\\
Can	&	64.1 	&	71.5 	&	\textbf{72.9} 	&	91.2 	&	98.1 	&	\textbf{98.4} 	\\
Cat	&	62.6 	&	65.5 	&	\textbf{67.1} 	&	\textbf{98.8} 	&	98.7 	&	\textbf{98.8} 	\\
Driller	&	74.4 	&	81.5 	&	\textbf{82.6} 	&	80.9 	&	88.8 	&	\textbf{90.1}	\\
Duck	&	44.3 	&	55.7 	&	\textbf{59.0}	&	92.2 	&	97.0 	&	\textbf{97.3} 	\\
Eggbox	&	57.8 	&	74.6 	&	\textbf{77.0} 	&	91.0 	&	95.2 	&	\textbf{95.5} 	\\
Glue	&	41.2 	&	83.4 	&\textbf{85.1} 	&	92.3 	&	97.8 	&	\textbf{98.5} 	\\
Holepuncher	&	67.2 	&	73.9 	&	\textbf{75.7} 	&	95.3 	&	97.8 	&	\textbf{98.2} 	\\
Iron	&	84.7 	&	87.8 	&\textbf{89.3}	&	84.8 	&	88.5 	&	\textbf{89.9} 	\\
Lamp	&	76.5 	&	80.7 	&	\textbf{83.1} 	&	75.8 	&	86.5 	&	\textbf{88.1} 	\\
Phone	&	54.0 	&	62.1 	&	\textbf{65.6}	&	85.3 	&	90.6 	&	\textbf{91.0} 	\\
\midrule
Average	&	62.7 	&	72.4 	&	\textbf{74.2} 	&	89.3 	&	94.6 	&	\textbf{95.3}	\\
\bottomrule
\end{tabular} \label{tab1}
\end{table}
\unskip
\begin{table}[H]
\caption{Pose estimation results of CorrNet (ours) and RANSAC \cite{ref11}
based method on an Occluded LineMOD dataset \cite{ref6}. The best results for each term are shown in bold.}
\centering
\scalebox{0.95}[0.95]{\begin{tabular}{l*{7}{c}}
\toprule
\multirow{2}*{\textbf{Sequence}\vspace{-4pt}} & \multicolumn{3}{c}{\textbf{AD{D|I}}} & \multicolumn{3}{c}{\textbf{2D Reprojection Error}}  \\
\cmidrule{2-7}
& \textbf{RANSAC \cite{ref11}} & \textbf{CorrNet w/o CC} & \textbf{CorrNet}& \textbf{RANSAC \cite{ref11}} & \textbf{CorrNet w/o CC} & \textbf{CorrNet}\\
\midrule

Ape	&	16.5 	&	17.0 	&	\textbf{17.3} 	&	64.7 	&	66.5 	&	\textbf{68.6}	\\
Can	&	42.5 	&	45.8 	&	\textbf{49.2} 	&	53.0 	&	61.7 	&	\textbf{64.9} 	\\
Cat	&	2.8 	&	2.9 	&	\textbf{3.0} 	&	47.9 	&	51.4 	&	\textbf{53.3} 	\\
Driller	&	47.1 	&	54.6 	&	\textbf{57.7} 	&	35.1 	&	48.5 	&	\textbf{55.0} 	\\
Duck	&	11.0 	&	12.1 	&	\textbf{13.2} 	&	36.1 	&	39.5 	&	\textbf{47.3} 	\\
Eggbox	&	24.7 	&	24.9 	&	\textbf{25.0} 	&	10.3 	&	10.3 	&	\textbf{10.4} 	\\
Glue	&	39.5 	&	39.7 	&	\textbf{39.9} 	&	44.9 	&	51.7 	&	\textbf{53.4} 	\\
Holepuncher	&	\textbf{21.9} 	&	\textbf{21.9 }	&	\textbf{21.9} 	&	52.9 	&	56.6 	&	\textbf{57.6} 	\\
\midrule
Average	&	25.8 	&	27.4 	&	\textbf{28.4} 	&	43.1 	&	48.3 	&	\textbf{51.3} 	\\

\bottomrule
\end{tabular} }\label{tab2}
\end{table}

\subsection{Results from the Full Pipeline}
We now evaluate our full pipeline on two datasets with serve occlusion, namely Occluded LineMOD dataset \cite{ref6}, and YCB-Video dataset \cite{ref19}. For comparison purposes, we have employed two state-of-the-art methods, that is, PoseCNN \cite{ref19} and DeepHMap \cite{ref11}. Note that all methods in the evaluation section take only RGB images as input. Especially for the YCB-Video dataset, the area under the accuracy-threshold curve (AUC) \cite{ref19} is utilized as an additional metric.

\begin{figure}[H]
\centering
\includegraphics[width=15 cm]{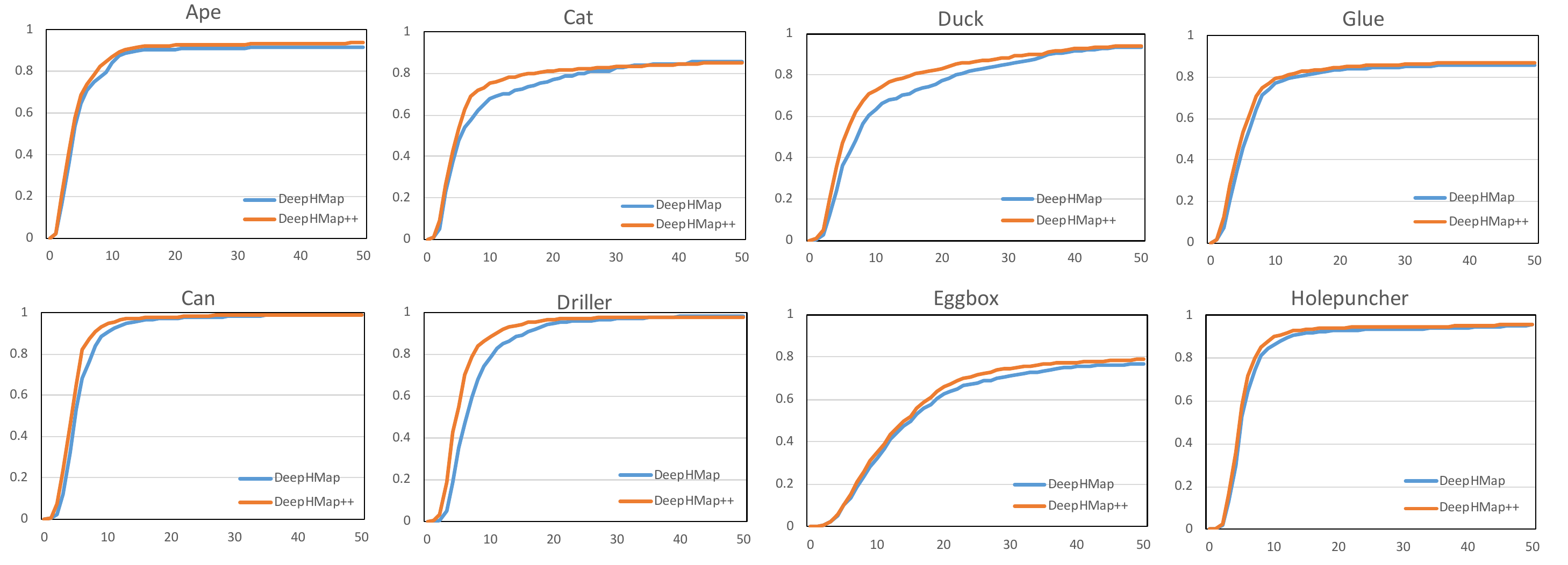}
\caption{Evaluations on an Occluded LineMOD dataset \cite{ref6}. The curve represents accuracy vs. pixel threshold in a 2D reprojection error metric. The vertical axis denotes a fraction of correctly estimated scenes. The horizontal axis denotes pixel threshold.} \label{fig8}
\end{figure}

As depicted in Figure \ref{fig8}, a more complete comparison between DeepHMap and DeepHMap++ is given. For all eight sequences from the Occluded LineMOD dataset \cite{ref6}, DeepHMap++ steadily achieves better results than DeepHMap under different pixel thresholds. With the 2D reprojection error metric, a smaller pixel threshold means more accurate estimation. It is not unusual to find that the boosting of DeepHMap++ is more obvious under a low-threshold phase that ranges from $ 0~px $ to $ 30~px $. This is because, when a test scene corresponds to a larger pixel threshold, it means that the estimated pose deviates significantly from the ground truth. Dealing with such challenging scenes is very difficult for both projection grouping module and correspondence evaluation module. Thus, the improvement of DeepHMap++ becomes limited when pixel threshold reaches a high level that is bigger than $ 30~px $.

\unskip
\begin{table}[H]
\caption{Comparisons with state-of-the-art methods on YCB-Video dataset \cite{ref19}. We report the AUC scores, AD{D|I} and 2D reprojection error for the 21 image sequences of YCB-Video dataset. The best results for each term are shown in bold.}
\centering
\scalebox{0.9}[0.9]{\begin{tabular}{l*{10}{c}}
\toprule
\multirow{2}*{\textbf{Sequence}} & \multicolumn{3}{c}{\textbf{PoseCNN \cite{ref19}}} & \multicolumn{3}{c}{\textbf{ DeepHMap \cite{ref11}}} & \multicolumn{3}{c}{\textbf{ DeepHMap++}}\\
\cmidrule{2-10}
& \textbf{AUC} & \textbf{AD{D|I}} & \textbf{2D Repr.}& \textbf{AUC} & \textbf{AD{D|I}} & \textbf{2D Repr.} & \textbf{AUC} & \textbf{AD{D|I}} & \textbf{2D Repr.}\\
\midrule
002 master chef can	&	50.1 	&	3.6 	&	0.1 	&	68.5 	&	32.9 	&	9.9 	&	\textbf{75.8 }	&	\textbf{40.1} 	&	\textbf{20.1} 	\\
003 cracker box	&	52.9 	&	25.1 	&	0.1 	&	74.7 	&	62.6 	&	24.5 	&	\textbf{78.0} 	&	\textbf{69.5} 	&	\textbf{34.5} 	\\
004 sugar box	&	68.3 	&	40.3 	&	7.1 	&	74.9 	&	44.5 	&	47.0 	&	\textbf{76.5} 	&	\textbf{49.7} 	&	\textbf{58.9} 	\\
005 tomato soup can	&	66.1 	&	25.5 	&	5.2 	&	68.7 	&	31.1 	&	41.5 	&	\textbf{72.1 }	&	\textbf{36.1} 	&	\textbf{49.8} 	\\
006 mustard bottle	&\textbf{80.8 }	&	\textbf{61.9} 	&	6.4 	&	72.6 	&	42.0 	&	42.3 	&	78.9 	&	57.9 	&	\textbf{60.1 }	\\
007 tuna fish can	&	\textbf{70.6} 	&	\textbf{11.4} 	&	3.0 	&	38.2 	&	6.8 	&	7.1 	&	51.6 	&	9.8 	&	\textbf{19.5 }	\\
008 pudding box	&	62.2 	&	14.5 	&	5.1 	&	82.9 	&	58.4 	&	43.9 	&	\textbf{85.6 }	&	\textbf{67.2} 	&	\textbf{56.8} 	\\
009 gelatin box	&	74.8 	&	12.1 	&	15.8 	&	82.8 	&	42.5 	&	62.1 	&	\textbf{86.7} 	&	\textbf{59.1} 	&	\textbf{76.8} 	\\
010 potted meat can	&	59.5 	&	18.9 	&	23.1 	&	66.8 	&	37.6 	&	38.5 	&	\textbf{70.1 }	&	\textbf{42.0 }	&	\textbf{42.3} 	\\
011 banana	&	\textbf{72.1 }	&	\textbf{30.3} 	&	0.3 	&	44.9 	&	16.8 	&	8.2 	&	47.9 	&	19.3 	&	\textbf{10.5} 	\\
019 pitcher base	&	53.1 	&	15.6 	&	0.0 	&	70.3 	&	57.2 	&	15.9 	&	\textbf{71.8 }	&	\textbf{58.5 }	&	\textbf{19.8} 	\\
021 bleach cleanser	&	50.2 	&	21.2 	&	1.2 	&	67.1 	&	65.3 	&	12.1 	&	\textbf{69.1} 	&	\textbf{69.4} 	&	\textbf{18.5 }	\\
024 bowl	&	\textbf{69.8} 	&	12.1 	&	4.4 	&	58.6 	&	25.6 	&	16.0 	&	60.2 	&	\textbf{27.7} 	&	\textbf{18.1} 	\\
025 mug	&	\textbf{58.4} 	&	5.2 	&	0.8 	&	38.0 	&	11.6 	&	20.3 	&	43.4 	&	\textbf{12.9 }	&	\textbf{26.3} 	\\
035 power drill	&	55.2 	&	29.9 	&	3.3 	&	72.6 	&	46.1 	&	40.9 	&	\textbf{76.8 }	&	\textbf{51.8 }	&	\textbf{50.1} 	\\
036 wood block	&	\textbf{61.8} 	&	10.7 	&	0.0 	&	57.7 	&	34.3 	&	2.5 	&	61.3 	&	\textbf{35.7} 	&	\textbf{2.8} 	\\
037 scissors	&	35.3 	&	\textbf{2.2} 	&	0.0 	&	30.9 	&	0.0 	&	0.0 	&	\textbf{42.9} 	&	2.1 	&	\textbf{6.7 }	\\
040 large marker	&	\textbf{58.1} 	&	3.4 	&	\textbf{1.4} 	&	46.2 	&	3.2 	&	0.0 	&	47.6 	&	\textbf{3.6} 	&	0.8 	\\
051 large clamp	&	\textbf{50.1} 	&	\textbf{28.5} 	&	0.3 	&	42.4 	&	10.8 	&	0.0 	&	44.1 	&	11.2 	&	\textbf{8.7 }	\\
052 extra large clamp	&	46.5 	&	19.6 	&	0.6 	&	48.1 	&	29.6 	&	0.0 	&	\textbf{51.9} 	&	\textbf{30.9} 	&	\textbf{0.8} 	\\
061 foam brick	&	\textbf{85.9} 	&	54.5 	&	0.0 	&	82.7 	&	51.7 	&	52.4 	&	84.1 	&	\textbf{55.4} 	&	\textbf{59.7 }	\\
\midrule
Average	&	61.0 	&	21.3 	&	3.7 	&	61.4 	&	33.8 	&	23.1 	&	\textbf{65.5} 	&	\textbf{38.6 }	&	\textbf{30.6} 	\\
\bottomrule
\end{tabular}} \label{tab3}
\end{table}

\begin{figure}[H]
\centering
\includegraphics[width=12 cm]{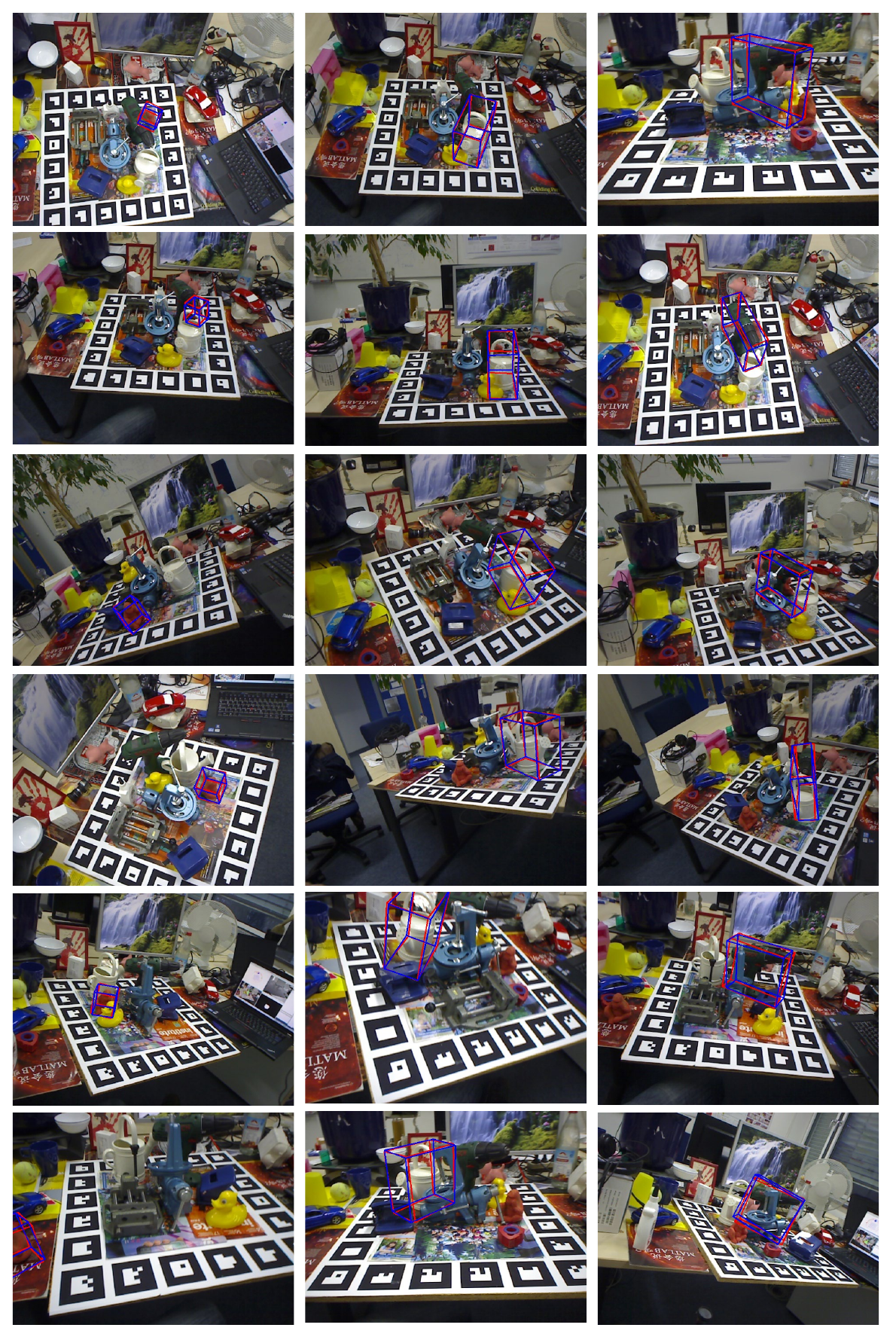}
\caption{Estimated 6D pose on an Occluded LineMOD dataset \cite{ref6}. The red and blue bounding boxes denote the ground truth and results estimated by DeepHMap++, respectively. The left column is the results of Ape sequence. The middle column is the results from Can sequence. The right column is the results from Driller sequence.} \label{fig9}
\end{figure}

\begin{figure}[H]
\centering
\includegraphics[width=13.5 cm]{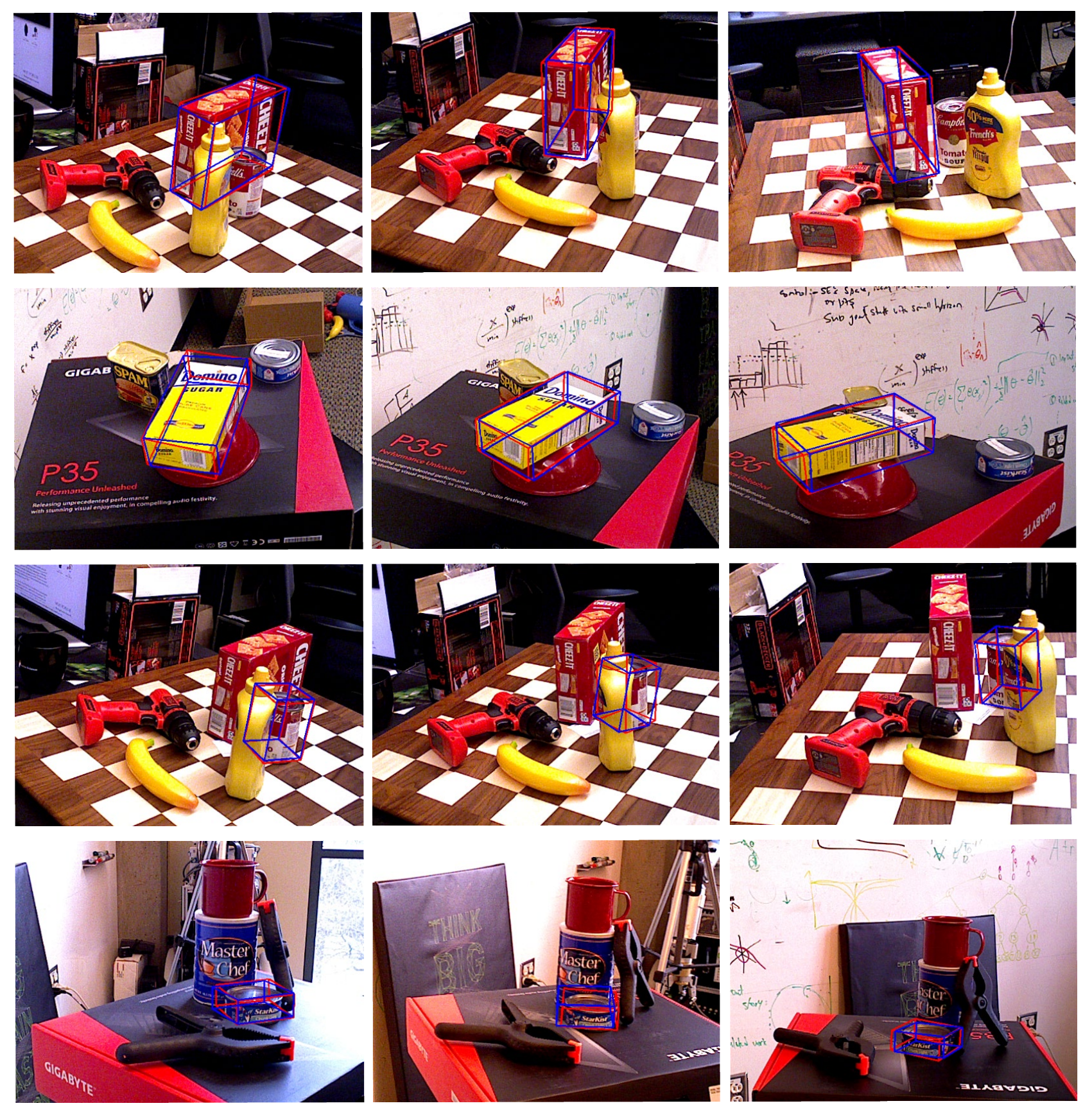}
\caption{Estimated 6D pose on YCB-Video dataset \cite{ref19}. The ground truth is shown in red, and estimated results with DeepHMap++ are shown in blue. The four rows (from up to down) correspond to test images from 003\_cracker\_box sequence, 004\_sugar\_box sequence, 005\_tomato\_soup\_can  sequence and 007\_tuna\_fish\_can sequence, respectively.} \label{fig10}
\end{figure}

Comparisons between DeepHMap++, DeepHMap and PoseCNN are listed in Table \ref{tab3}. For all object sequences from YCB-Video dataset, DeepHMap++ consistently improves DeepHMap, which utilizes RANSAC based correspondence sampling and $ max $ function based projection grouping in three different metrics. For another baseline \cite{ref19}, semantic labeling and object center are jointly employed intermediate cues. However, the entire image is directly cast into CNN for building the mapping from image space to object center. This holistic scheme is more sensitive to foreground occlusion than both DeepHMap and DeepHMap++ using local input. Local feature input plus a specially designed back-end ensures that DeepHMap++ achieves best results over most of the entries. We also show some qualitative results on both datasets in Figures \ref{fig9} and \ref{fig10}, respectively.

\subsection{Runtime Analysis}
Our current implementation is written in python on an Ubuntu machine with an intel E5-2640 CPU (intel, Santa Clara, CA, USA) 
and NVIDIA Geforce GTX1080Ti GPU (NVIDIA, Santa Clara, CA, USA). 
The network part is built on a large-scale machine learning library-tensorflow \cite{ref50}. In~the first stage, the projection heatmap prediction takes 80 ms for 64~patches. Parallel processing can significantly reduce the prediction time to 20 ms. Benefiting from the simple architecture of projection grouping module, it only takes 3 ms to complete this subtask. Subsequent~ correspondence evaluation network takes 20 ms to assign correspondence weights and compute the full DoF pose parameters. For~a  640 px $\times$ 480 px image, it takes about 130 ms for pose detection via a sliding window fashion and goes down to 60 ms with a parallel trick. The main runtime statistics are listed in Table \ref{tab4}.

\begin{table}[H]
\caption{The runtime statistics of different subparts in the complete pipeline. }
\centering
\begin{tabular}{cll}
\toprule
& \textbf{Processing Step} & \textbf{Time}\\
\midrule
\multirow{2}{*}{First stage} & Projection heatmap predictions        &  80  ms \\
& Projection heatmap predictions (in parallel)                       &  20 ms   \\
\multirow{2}{*}{Second stage} & Projection grouping                  &  3  ms   \\
& Correspondence evaluation                              &  20  ms   \\
Full pipeline        &                                   & $ 80 + 3 + 20 = 103$ ms  \\
Full pipeline (in parallel)     &                       & $ 20 + 3 + 20= 43$ ms  \\
\bottomrule
\end{tabular} \label{tab4}
\end{table}

\section{Conclusions and Future Work} \label{sec5}

We have improved the back-end of a two-stage pipeline to recover the 6D pose of rigid objects under challenging scenes. With a simple fully connected module, the projection ambiguity can be better addressed than the one-size-fits-all strategy in DeepHMaps. The proposed projection grouping module learns correlation constraints of different BBCs and reduces the number of false projection selections. A~corresponding-evaluation network is then employed to achieve weighted correspondences, as~opposed to RANSAC based strategy. The above mentioned efforts have enabled the proposed method to outperform state-of-the-art solutions on three public benchmarks. Meanwhile, these refinements don’t introduce too much computing burden, which indicates the great potential of our method in real-time applications.

In the future, an interesting direction is to add a branch on the backbone for object segmentation. This branch can provide additional regularization to some extent. Another line is to fuse the improved two-stage approach into a pose tracking framework. In a standard pipeline of pose tracking, pose parameters from the previous frame can be reused to replace the pose detection step in DeepHMap. In~addition, tuning the architecture of a network to achieve an end-to-end training is beneficial to final~results.

\vspace{6pt}



\authorcontributions{M.F. conceived of and designed the experiments. M.F. performed the experiments. M.F. analyzed the data. W.Z. supervised this work.}

\funding{This research was funded by the National Science Foundation of China Grant No. 51505470. 
}

\acknowledgments{The authors would like to thank Stefan Hinterstoisser (Google), Eric Brachmann (Heidelberg University) and Xiang Yu (NVIDIA) 
for making related datasets freely available to the public.}

\conflictsofinterest{The authors declare no conflict of interest.}






\reftitle{References}





\end{document}